\documentclass[sn-mathphys-num]{sn-jnl}

\usepackage{graphicx}%
\usepackage{multirow}%
\usepackage{amsmath,amssymb,amsfonts}%
\usepackage{amsthm}%
\usepackage{mathrsfs}%
\usepackage[title]{appendix}%
\usepackage{xcolor}%
\usepackage{textcomp}%
\usepackage{manyfoot}%
\usepackage{booktabs}%
\usepackage{algorithm}%
\usepackage{algorithmicx}%
\usepackage{algpseudocode}%
\usepackage{listings}%

\usepackage{indentfirst}
\usepackage[utf8]{inputenc}

\usepackage{makecell}
\usepackage{subfigure}

\begin{document}

\title[Article Title]{PdfTable: A Unified Toolkit for Deep Learning-Based Table Extraction}


\author*[1]{\fnm{Lei} \sur{Sheng}}\email{xuanfeng1992@whut.edu.cn}

\author[2]{\fnm{Shuai-Shuai} \sur{Xu}}\email{sa517432@mail.ustc.edu.cn}


\affil*[1]{\orgdiv{Automated institute}, \orgname{Wuhan University of Technology}, \orgaddress{\street{122 Luoshi Road}, \city{Wuhan}, \postcode{430070}, \state{Hubei}, \country{China}}}

\affil[2]{\orgdiv{School of Software}, \orgname{University of Science and Technology of China}, \orgaddress{\street{No.96, JinZhai Road Baohe District}, \city{Hefei}, \postcode{230026}, \state{Anhui}, \country{China}}}


\abstract{
    Currently, a substantial volume of document data exists in an unstructured format, encompassing Portable Document Format (PDF) files and images. Extracting information from these documents presents formidable challenges due to diverse table styles, complex forms, and the inclusion of different languages. Several open-source toolkits, such as Camelot, Plumb a PDF (pdfnumber), and Paddle Paddle Structure V2 (PP-StructureV2), have been developed to facilitate table extraction from PDFs or images. However, each toolkit has its limitations. Camelot and pdfnumber can solely extract tables from digital PDFs and cannot handle image-based PDFs and pictures. On the other hand, PP-StructureV2 can comprehensively extract image-based PDFs and tables from pictures. Nevertheless, it lacks the ability to differentiate between diverse application scenarios, such as wired tables and wireless tables, digital PDFs, and image-based PDFs. To address these issues, we have introduced the PDF table extraction (PdfTable) toolkit. This toolkit integrates numerous open-source models, including seven table recognition models, four Optical character recognition (OCR) recognition tools, and three layout analysis models. By refining the PDF table extraction process, PdfTable achieves adaptability across various application scenarios. We substantiate the efficacy of the PdfTable toolkit through verification on a self-labeled wired table dataset and the open-source wireless Publicly Table Reconition Dataset (PubTabNet). The PdfTable code will available on Github: \url{https://github.com/CycloneBoy/pdf_table}.
}




\maketitle
\section{Introduction}
\label{sec:introduction}
Portable Document Format (PDF\footnote{\url{https://en.wikipedia.org/wiki/PDF}}), a file format used to present documents in a hardware- and software-independent manner. It finds widespread application across various domains, including academic papers and financial report documents. With the rapid development of document digitization, the automated extraction of information from PDFs has gained paramount significance. Consequently, several tools have emerged to facilitate the conversion of PDFs into easily parseable HTML formats. However, the intricate structures and diverse styles of tables, coupled with the potential inclusion of different languages in table contents, present persistent challenges in table structure recognition(TSR) during the parsing of PDF documents. In response to this challenge, diverse methods have been proposed to address the complexities associated with TSR.

Several toolkits are available to directly extract tables from PDFs, including Tabula\footnote{\url{https://github.com/tabulapdf/tabula}}, Camelot\footnote{\url{https://github.com/camelot-dev/camelot}} and pdfnumber\footnote{\url{https://github.com/jsvine/pdfplumber}}. Tabula, Camelot, and pdfnumber primarily employ rule-based methods for table extraction in digital PDFs, demonstrating inaccuracies when confronted with tables featuring complex cross-line and cross-column styles. With the rapid development of deep learning, early researchers proposed models such as DeepDeSRT\cite{DeepDeSRT_2017} ,TableNet\cite{paliwal_tablenet_2020} and SEM\cite{zhang_split_2022} to address table extraction challenges in image-based documents. However, due to a scarcity of extensively annotated datasets, the outcomes were less than satisfactory. Recent years have witnessed the introduction of diverse TSR datasets, such as SciTSR\cite{chi2019complicated}, TableBank\cite{li2020tablebank}, PubtabNet\cite{zhong2020PubtabNet} , PubTables-1M\cite{Smock2021PubTables1MTC}, and WTW\cite{long2021WTW}. Models like CascadeTabNet\cite{prasad_cascadetabnet_2020}, EDD\cite{zhong2020PubtabNet}, LGPMA\cite{qiao2021lgpma}, TSRFormer\cite{Chixiang2023TSRFormer}, Cycle-CenterNet\cite{long2021WTW}, LORE\cite{Hangdi2023LORE}, etc., trained on these datasets have demonstrated proficient table parsing results. Despite successful parsing in generalized table scenarios, these models encounter challenges when applied to real-world scenarios.

While the existing table parsing algorithm performs admirably, there remains a deficiency in open-source tools designed for end-to-end PDF table extraction to address diverse table extraction tasks in practical applications.
Baidu's recent open source PP-StructureV2\cite{ppstructurev2_li2022pp} toolkit, employing the SLANet table structure recognition model in conjunction with PaddleOCR\cite{du2020ppocr}, has garnered widespread user appreciation for achieving end-to-end table recognition and extraction. Nevertheless, there are notable areas for optimization: (1) the end-to-end table extract process lacks sufficient subdivision, such as the differentiation between wired and wireless tables, and the extraction of text from digital PDFs versus image-based PDFs; (2) Each functional module in the recognition process supports a limited number of models, for instance: two layout analysis models, three table recognition models and one OCR text recognition model; (3) Open source table recognition models commonly employ different frameworks and dependent environments, posing challenges for debugging and reproducibility within a unified environment.

In addressing the aforementioned challenges, we present a novel end-to-end PDF extraction table toolkit called PdfTable. Initially, we partition the table recognition process into distinct modules, including data preprocessing, layout analysis, table structure recognition, table content extraction and upper-layer application. Then different open source algorithms and toolkits are integrated for different modules. Diverse open-source algorithms and toolkits are integrated for each module, with uniform coding implemented in Pytorch\cite{paszke2019pytorch} to streamline debugging and model integration. Presently, the toolkit encompasses seven table structure recognition algorithms, three layout analysis algorithms, and four mainstream OCR recognition tools. 
Subsequently, we conduct end-to-end integration and optimization of the table recognition process, ultimately enabling the batch conversion of both digital and scanned PDF documents into HTML or WORD formats. PdfTable facilitates the direct extraction of PDF tables into Excel and supports numerous languages. To validate the toolkit's effectiveness, we annotated a small table dataset within the Chinese financial domain, comprising both digital and scanned PDFs. PdfTable demonstrated commendable performance on this dataset, affirming the efficacy of the toolkit. Concurrently, we evaluated the integration of four wireless table models on the PubtabNet\cite{zhong2020PubtabNet} wireless table dataset, with results attesting to the correctness of the model integration.

In summary, our primary contributions can be outlined as follows:
\begin{enumerate}
    \item Introduction of PdfTable, an end-to-end deep learning-based PDF table extraction toolkit, supporting the extraction of tables from both digital and scanned PDFs, encompassing wired and wireless table extraction.
    \item Integration of numerous open-source algorithms into our toolkit, encompassing seven table structure recognition models, four mainstream OCR recognition tools, and three layout analysis models. This integration provides users with a straightforward and user-friendly API.
    \item Conducted experiments to validate the efficacy of the PdfTable toolkit, utilizing a self-labeled small Chinese financial field wired table dataset and the wireless table dataset PubtabNet\cite{zhong2020PubtabNet}. The experimental results unequivocally demonstrate the effectiveness and correctness of our toolkit.
\end{enumerate}

\section{Related Work}
\label{sec:related_works}

\subsection{Document Layout Analysis}
\label{subsec:ly}
Document layout analysis is a basic pre-processing task for modern document understanding and digitization. It mainly divides documents into different regions, such as pictures, tables, text and formulas, which can be regarded as a sub-task of object detection. Presently mainstream methods include object detection-based models, segmentation-based models and GNN-based methods\cite{wei_paragraph2graph_2023}. DeepDeSRT\cite{DeepDeSRT_2017} pioneered the use of Faster R-CNN\cite{girshick2015fast} for table detection, achieving commendable results. With more and more layout analysis datasets ,such as: PubLayNet\cite{PubLayNet_2019}, TableBank\cite{li2020tablebank}, etc. and different object detection models such as Mask R-CNN\cite{he2017mask}, YOLO\cite{redmon2016you}, DETR\cite{carion2020end}, etc. proposed, the task of layout analysis has been further developed. Layout-parser\cite{shen2021layoutparser} is a unified toolkit for document image analysis based on deep learning, providing rich pre-trained models and user-friendly APIs. PP-StructureV2\cite{ppstructurev2_li2022pp} also provides a variety of English and Chinese layout analysis models trained on PP-YOLOv2\cite{ppdet2019} and PP-PicoDet\cite{yu2021pp} models.

\subsection{Table Structure Recognition}
\label{subsec:tsr}
Historically, early approaches to table recognition primarily employed rule-based and statistical machine learning methods, often limited by their dependence on the rigid rectangular layout of tables. Consequently, these methods could only effectively handle straightforward table structures or tables embedded in PDFs. In recent years, the landscape has shifted towards deep learning-based methods, demonstrating substantial improvements in accuracy compared to traditional approaches. Broadly categorized, these contemporary methods fall into three main groups: boundary extraction-based methods, image-to-markup generated methods, and graph-based methods.

\textbf{Boundary extraction based methods} These methods employ object detection or semantic segmentation algorithms to initially identify the rows and columns of the table. Subsequently, the cells of the table are determined through cross-combination of the identified rows and columns. DeepDeSRT\cite{DeepDeSRT_2017} and TableNet\cite{paliwal_tablenet_2020} leverage Fully Convolutional Network (FCN)-based semantic segmentation model for TSR analysis. However, the basic FCN faces challenges in accurately recognizing numerous blank tables due to its limited receptive field. To address this limitation, subsequent researchers proposed enhancements \cite{DeepTabStR_2019, zhang_split_2022,prasad_cascadetabnet_2020}. SEM\cite{zhang_split_2022} stands out by integrating visual and textual information through three independent modules—splitter, embedder, and merge—enabling the extraction of both simple and complex tables. RobusTabNet\cite{Chixiang2023TSRFormer} proposed a new method of splitting and merging TSR using spatial Convolutional Neural Network (CNN) module, which can effectively identify tables with a large number of blanks and distortions.

\textbf{Image-to-markup generation based methods} These methods transform the table recognition task into an image-to-markup generation task, directly generating markup (HTML or LaTeX) to represent the table structure. Leveraging a substantial volume of labeled table data extracted from existing PDFs, web pages, or LaTeX papers through rules or semi-supervised methods, researchers have proposed many benchmark datasets TABLE2LATEX\cite{icdar2019_table}, Tablebank\cite{li2020tablebank}, PubtabNet\cite{zhong2020PubtabNet}. Additionally, they have organized related competitions, including ICDAR2019\cite{icdar2019_table}, ICDAR2021\cite{noauthor_icdar_2021}, significantly fostering the rapid development of TSR. TableMaster\cite{ye2021TableMaster} directly predicts HTML and text box regression based on MASTER\cite{LU2021107980}, achieving the best results on the PubtabNet benchmark dataset. SLANet\cite{ppstructurev2_li2022pp} uses PP-LCNet\cite{cui_pp-lcnet_2021} and a series of optimization strategies to make model inference efficient on the CPU. MTL-TabNet\cite{visapp23MTLTabNet} proposes an end-to-end TSR model that uses a multi-task learning method to directly solve table structure recognition and table content recognition with one model. OTSL\cite{OTSL_2023} proposes a novel method for marking tables, utilizing only five tokens to represent the table structure, thereby reducing the inference time of the Image2seq method by approximately half while enhancing model accuracy. Only five tokens can be used to represent the table structure, which can shorten the inference time of the Image2seq method by about half while improving model accuracy.

\textbf{Graph based methods} These methods treat table cells or cell contents as nodes in a graph, employing graph neural networks(GNN) to predict whether these nodes belong to the same group. GraphTSR\cite{chi2019complicated} takes table cells as input, and then uses GNN to predict the relationship between table cells to predict the table structure, achieving good results on the SciTSR\cite{chi2019complicated} data set. TGRNet\cite{TGRNet_2021} proposes an end-to-end table graph reconstruction network to perform table structure recognition by simultaneously predicting the physical and logical positions of table cells.

\subsection{Optical Character Recognition}
\label{subsec:ocr}

Table content recognition is also a crucial phase in the table recognition process. Tables in digital PDFs can directly read text coordinates and content, while scanned PDFs usually require an OCR model to extract text. OCR is currently divided into two primary tasks: text detection and text recognition, each optimized independently. Additionally, there are also end-to-end recognition models. Since OCR has a wide range of applications, it has received widespread attention from researchers and industries, leading to the proposal of numerous models. Notably, the DB\cite{liao2020real} detection model and CRNN\cite{shi2016end} recognition model stand out as widely adopted combinations. Several readily available open-source toolkits (such as PaddleOCR\cite{du2020ppocr}, EasyOCR\footnote{\url{https://github.com/JaidedAI/EasyOCR}}, TesseractOCR\footnote{\url{https://github.com/tesseract-ocr/tesseract}}, MMOCR\footnote{\url{https://github.com/open-mmlab/mmocr}}, and duguangOCR\footnote{\url{https://github.com/AlibabaResearch/AdvancedLiterateMachinery}}) and commercial APIs (Amazon Textract\footnote{\url{https://aws.amazon.com/textract/}}, Google Document ai\footnote{\url{https://cloud.google.com/document-ai}}, BaiDu OCR\footnote{\url{https://ai.baidu.com/tech/ocr}}) offer fundamental OCR capabilities. The majority of these open-source toolkits provide the latest OCR algorithms and pre-trained models, facilitating convenient direct use or fine-tuning. Due to the distinct nature of document OCR, it is readily identifiable, and existing open-source toolkits can effectively fulfill most requirements.

\subsection{PDF To HTML}
\label{subsec:pdf2html}
The conversion of PDF to machine-readable HTML format holds significant implications. For instance, it can enhance accessibility for individuals who are blind or visually impaired\cite{fayyaz_enhancing_2023} and contribute to the improved retrieval and dissemination of academic papers\cite{wang_scia11y_2021,ahuja_analyzing_2023}. While several off-the-shelf systems exist for direct PDF-to-HTML conversion, they exhibit limitations\cite{shigarov_tabbypdf_2018,pr_dexter_2022,rao_tableparser_2022}. Notably, \cite{wang_scia11y_2021,ahuja_analyzing_2023} lacks table parsing functionality, converting PDF tables into images for display only. \cite{namysl_flexible_2021} relies on hand-designed rules for table extraction, demonstrating poor generalization. TableParser\cite{rao_tableparser_2022} is a model trained based on a weakly supervised dataset constructed from spreadsheets and cannot parse wireless tables or deformed tables. Pdf2htmlEX\footnote{\url{https://github.com/pdf2htmlEX/pdf2htmlEX}} exclusively converts digital PDFs, failing to convert tables into HTML format. The end-to-end model proposed by Nougat\cite{blecher_nougat_2023}, based on a visual Transformer, excels in converting academic papers into LaTeX format.
However, its end-to-end nature necessitates data collection for retraining when dealing with PDFs in different languages or structures, imposing certain limitations on its versatility. Despite the effectiveness of these systems in specific scenarios, a unified PDF-to-HTML conversion tool supporting diverse languages and document types remains elusive.

\begin{figure}[t]
	\begin{center}
		  \includegraphics[width=0.95\textwidth]{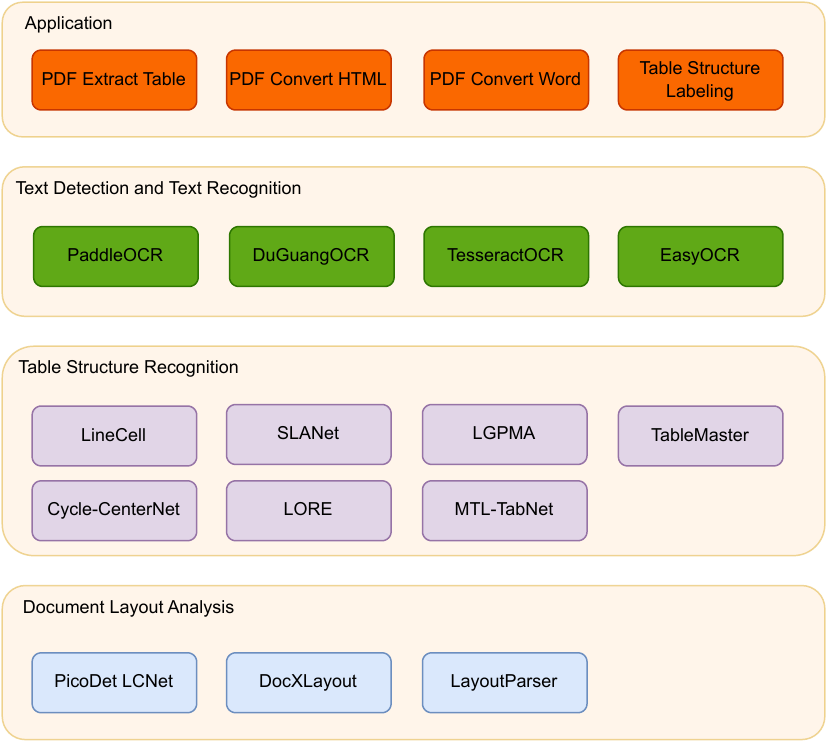}
		  \caption{System overview of PdfTable}
		  \label{figure:system_overview} 
	\end{center}
\end{figure}
\section{Design and implementation of PdfTable library}
\label{sec:methodology}

\subsection{System Overview}
\label{subsec:system_overview}

The system overview of PdfTable is illustrated in Figure \ref{figure:system_overview}. The core of the entire system is to provide table parsing related algorithms, which is mainly composed of four modules. The layout analysis module locates tables and images; the table structure recognition module parses table structures; text detection and recognitionc module identifies textual content; the application module primarily handles the conversion of all recognition results into various types. We have standardized the algorithm interface for each module, allowing flexible switching based on the model name to facilitate user utilization. Since different algorithms rely on different environments and frameworks, we use the Pytorch\cite{paszke2019pytorch} framework to reconstruct part of the model, eliminating unnecessary dependency packages.

\subsection{PdfTable Parse Pipline}
\label{subsec:parse_pipline}

\begin{figure}
	\begin{center}
		  \includegraphics[width=0.95\textwidth]{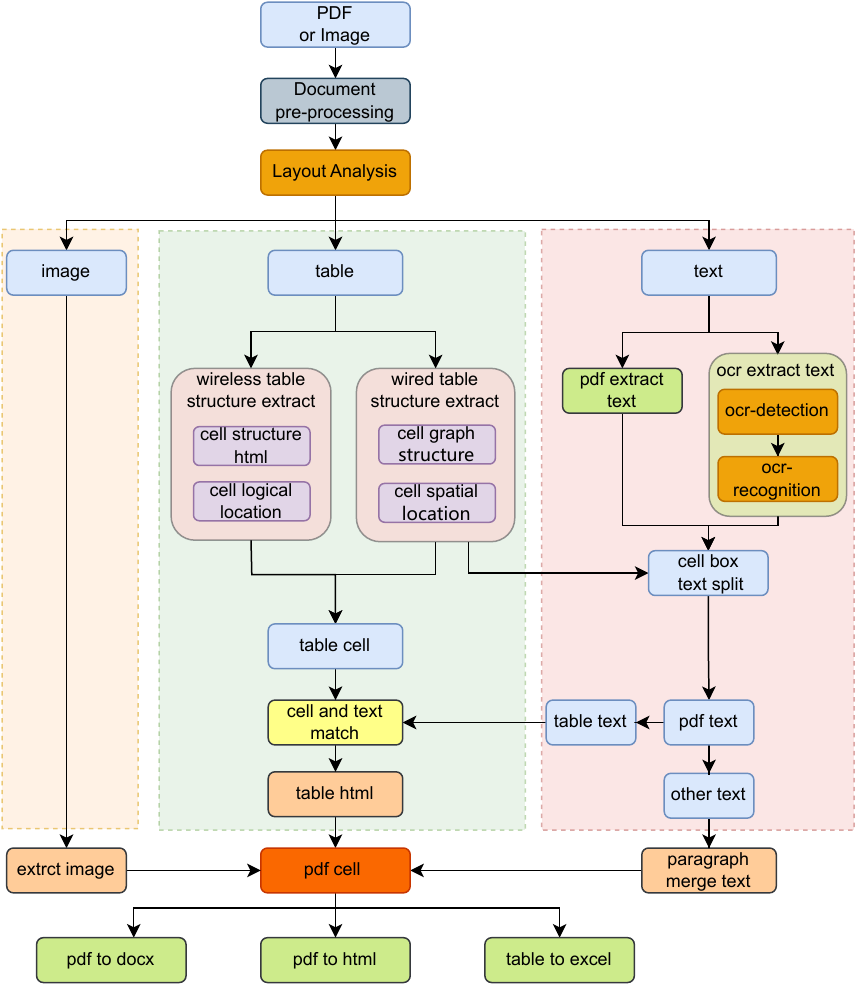}
		  \caption{table processing pipline}
		  \label{figure:table_pipline} 
	\end{center}
\end{figure}

The framework of table recognition is illustrated in Figure \ref{figure:table_pipline}. Firstly, the input image or PDF document is preprocessed through the document preprocessing module, such as network file download, PDF conversion to image, image orientation correction, etc. Subsequently, the layout analysis module divides the image into distinct regions (e.g., pictures, tables, and text) to facilitate subsequent individual processing. The image area is sent to the image extraction module for extraction. The table area distinguishes whether it is a wired table or a wireless table through rules, and then extracts the table structure through the TSR algorithm. For the text area, extraction is performed based on the document type. In the case of digital PDFs, text is directly extracted from the PDF, while OCR is employed for scanned PDFs or images to identify the corresponding text. Next, the text in the table area is matched with the table structure to generate table HTML, and other text is consolidated into paragraphs through the paragraph merging module. Ultimately, the recognized pictures, tables, and text paragraphs are output into distinct files according to the specified output format requirements.

\subsection{Module Design}
\label{subsec:module_design}

\subsubsection{Input preprocessing} 
\label{subsubsec:preprocess}
\begin{figure}
	\begin{center}
		  \includegraphics[width=0.5\textwidth]{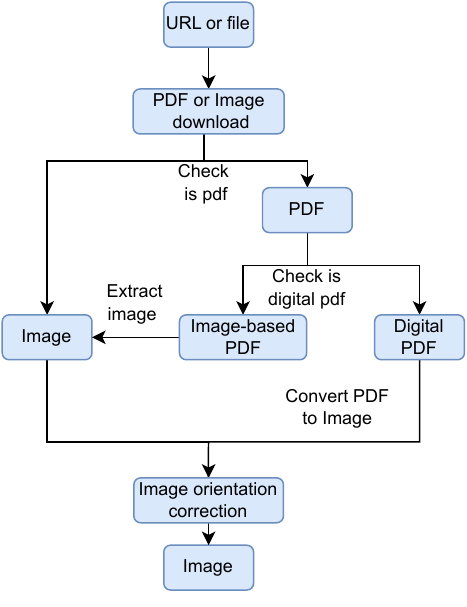}
		  \caption{table preprocess}
		  \label{figure:table_preprocess} 
	\end{center}
\end{figure}

The input preprocessing module primarily preprocesses PDFs or images to facilitate extraction of subsequent algorithm models. The processing flow chart is shown in Figure \ref{figure:table_preprocess}. Initially,the determination is made regarding the necessity of downloading the input file. If the input file is a PDF, it is requisite to split the PDF file into individual pages and convert them into images. For digital PDF, the Ghostscript\footnote{\url{https://www.ghostscript.com/}} tool is used to convert it into images, whereas for image PDFs, direct image extraction is conducted. Since the current document processing algorithm mainly processes documents with with a 0-degree orientation, the extraction outcome for rotated document information is suboptimal.  Consequently, the orientation of the input document must be rectified before the subsequent processing stages. The image orientation correction module incorporates the document orientation classification algorithm \footnote{ \url{https://www.ghostscript.com/}} (output categories: 0,90,180,270) and the text orientation classification algorithm (output categories: 0,180) to execute rotation correction based on the document orientation (rotation directions: 0,90,180,270). Concurrently, rules are applied for small-angle rotation correction (rotation angle: -45 - 45 degrees) on documents tilted at a slight angle, ultimately aligning the image for processing to roughly 0 degrees. The pre-processing module has significantly enhanced our recognition efficacy on irregular documents.

\subsubsection{Layout analysis module} 
\label{subsubsec:layout}
The task of layout analysis is to divide the areas in the document image according to categories (e.g.,text, images, tables, formulas). Currently, mainstream object detection-based models have demonstrated commendable performance across various benchmark datasets. In PdfTable, we have incorporated two lightweight layout analysis models, PP-picodet\cite{yu2021pp} and DocxLayout\footnote{\url{https://github.com/AlibabaResearch/AdvancedLiterateMachinery/tree/main/DocumentUnderstanding/DocXLayout}}, and the LayoutParser\cite{shen2021layoutparser} toolkit. PP-picodet is a lightweight target detection backbone model based on PaddleDetection\footnote{\url{https://github.com/PaddlePaddle/PaddleDetection}}, and ppstructure\cite{ppstructurev2_li2022pp} extends its capabilities to Chinese and English layout analysis models and table detection models. We convert them into a pytorch model. DocxLayout is a layout analysis model based on the DLA-34\cite{yu2018dla34} backbone network provided by Alibaba Research. The LayoutParser\cite{shen2021layoutparser} toolkit integrates a variety of layout analysis models based on different datasets. To enhance usability, we provide a standardized interface for invoking different models.

\subsubsection{Table Structure Recognition} 
\label{subsubsec:tsr}
Table borders are usually used for visual display of table structures and can also be used as an important basis for identifying table structures. The current mainstream method divides tables into two categories (wired tables and wireless tables), and designs TSR algorithms for processing different types of tables. In the TSR processing flow of PdfTable, an initial rule-based method is employed to categorize tables as either wired or wireless, followed by the application of specific algorithms to identify each table type. PdfTable currently integrates seven TSR algorithms, offering flexibility in configuration and utilization.

\textbf{Wired Table} Since wired tables have the obvious feature of borders, algorithms can be used to directly identify the borders of the table and then restore the table structure through post-processing. Traditional methods effectively handle most straightforward table scenes. We refer to camelot\footnote{\url{https://github.com/camelot-dev/camelot}} and Multi-TypeTD-TSR\cite{fischer2021multi} to implement the LineCell algorithm for extracting table cells based on OpenCV\cite{bradski2000opencv}. Firstly, we extract horizontal and vertical line segments, table areas, and intersections of line segments through a series of operations such as binarization, erosion, expansion, and contour search. Then we use line segment intersections and line segments to construct table cells. Finally, we use line segment relationships to merge across rows and columns cells. Despite their efficacy in simple scenes, traditional methods exhibit limited generalization due to their dependence on manually set rules. In contrast, contemporary approaches leverage deep learning techniques to identify table edges or cells. Therefore, PdfTable also integrates two latest TSR algorithms, Cycle-CenterNet\cite{long2021WTW} and LORE\cite{Hangdi2023LORE}. They adopt different methods to simultaneously predict the logical structure and physical structure of table cells, and then restore the structure of the table through simple post-processing operations, which can identify wired tables in real-world scenarios.

\textbf{Wireless Table} Wireless tables distinguished by the absence of table borders, present a more challenging identification task compared to wired tables. Presently, various methods employ image-to-sequence generation techniques, directly generating tags and text borders to represent the table structure. Subsequently, the table is reconstructed by aligning table cells with their respective content positions. We implemented four such algorithms: SLANet\cite{ppstructurev2_li2022pp}, LGPMA\cite{qiao2021lgpma}, TableMaster\cite{ye2021TableMaster}, and MTL-TabNet\cite{visapp23MTLTabNet}. The LORE\cite{Hangdi2023LORE} algorithm, while initially designed for recognizing wired tables, exhibits the capability to recognize wireless tables as well. This is achieved by predicting both the logical structure of the table and the physical borders of the table text.

\subsubsection{Text Extraction} 
\label{subsubsec:ocr}
In order to completely restore the table, it is also necessary to extract the content in the table cells (mainly text). The process of extracting table content in PdfTable is illustrated in the right part of Figure \ref{figure:table_pipline}, which is mainly divided into PDF text extraction and OCR text extraction. For digital PDF sources, the existing toolkit pdfminer.six\footnote{\url{https://github.com/pdfminer/pdfminer.six}} is utilized to directly extract text coordinates and content; otherwise, an OCR toolkit is employed. To accommodate multiple languages and diverse business scenarios, PdfTable integrates several mainstream OCR toolkits, including PaddleOCR\cite{du2020ppocr}, EasyOCR\footnote{\url{https://github.com/JaidedAI/EasyOCR}}, TesseractOCR\footnote{\url{https://github.com/tesseract-ocr/tesseract}} and duguangOCR\footnote{\url{https://github.com/AlibabaResearch/AdvancedLiterateMachinery}}. Combined with the table structure extracted previously, we split the text boxes across cells, and then match them with the table structure to generate the final table HTML. Text outside the table is merged into paragraphs to facilitate subsequent processing.

\subsubsection{Application} 
\label{subsubsec:app}
To address diverse application scenarios, we summarize the extracted table structure, text content and images and uniformly represent them into a PdfCell structure with coordinate positions and content. This approach facilitates the generation of diverse output formats. Currently, the applications implemented in PdfTable include: PDF to HTML, PDF to DOCX, and table to Excel. In the future, we will implement more different applications.

\section{Experiments}
\label{sec:experiments}

The primary objective of PdfTable is to streamline the extraction of tables from diverse PDF formats. The extraction of tables from PDFs poses challenges in practical applications due to variations in PDF types (digital or image-based), table categories (wired or wireless), and the presence of text in multiple languages (English, Chinese or other languages). A singular model proves insufficient for accommodating all business scenarios. PdfTable overcomes this limitation by integrating multiple models and allowing the selection of appropriate models for combined extraction based on distinct business types. Given that the extraction effectiveness of PdfTable depends on the chosen models and specific application contexts, a comprehensive evaluation is challenging.
Initial assessments were conducted on a common application scenario involving Chinese wired tables to validate the effectiveness of the PdfTable toolkit. Additionally, for wireless table recognition, an evaluation on the PubTabNet\cite{zhong2020PubtabNet} dataset was undertaken to verify the correctness of the integrated TSR algorithm.

\begin{table}[htbp]
	\centering
    \caption{Statistics of the datasets that we use in experiments. }
	\label{tab:test_dataset}
	\begin{tabular}{l|rr}
		\toprule
		\text{Test Sets}     & \text{Page}    & \text{Table}  \\
		\hline
		\text{Digital PDF}      & \text{2,589}   & \text{3,709}               \\
		\text{Image-based PDF}      & \text{2,192}   & \text{2,956}                   \\
		\hline
		\text{PubTabNet}      & \text{9,115}   & \text{9,115}                   \\
		\bottomrule
	\end{tabular}
	\footnotetext{}
\end{table}

\subsection{Datasets and evaluation metrics}
\label{subsec:dataset}
To assess PdfTable's capability in extracting wired tables, we curated a dataset comprising Chinese financial documents tables, encompassing both digital and image-based PDFs. This dataset comprises 4,781 pages and encompasses 6,665 tables. For the evaluation of wireless table extraction, we utilized the validation set from the extensively employed PubTabNet\cite{zhong2020PubtabNet} dataset. The details of the data set are shown in Table \ref{tab:test_dataset}.

We employ metrics such as Accuracy, Precision, Recall, F1-scores and TEDS-Struct\cite{raja_table_2020,ppstructurev2_li2022pp,Hangdi2023LORE} for evaluating table structure recognition. A table is deemed correctly recognized in Precision calculation when all its cells are accurately identified. TEDS-Struct is a modified variant of the tree edit distance-based similarity (TEDS)\cite{zhong2020PubtabNet} metric, which disregards the text content within table cells and exclusively evaluates the table structure.

 \begin{table}[htbp]
    \caption{Performance on the financial reporting dataset.}
    \label{tab:result_main}
	\begin{tabular*}{\textwidth}{c|c|ccc|c}    
		\toprule
        \text{Dataset}  & \text{Methods}  & \makecell{Precision\\(\%)}   & \makecell{Recall\\(\%)}  & \makecell{F1\\(\%)} & \makecell{TEDS-Struct \\ (\%)}  \\
		\hline
		 \multirow{2 }{*}{Digital PDF} & \text{LineCell} & \textbf{98.5}    & \textbf{98.2}     & \textbf{98.4}  & \textbf{99.5}   \\
		 & \text{LORE\cite[]{Hangdi2023LORE}}  & \text{90.5}    & \text{87.7}     & \text{89.1}  & \text{97.2}  \\
		 & \text{LORE$^*$\cite[]{Hangdi2023LORE}}  & \text{95.2}    & \text{93.2}     & \text{94.2}  & \text{98.4}  \\
         \midrule
		 \multirow{2 }{*}{Image-based PDF} & \text{LineCell} & \text{83.9}    & \textbf{84.7}     & \text{84.2}  & \text{94.7}   \\
		 & \text{LORE\cite[]{Hangdi2023LORE}}  & \text{80.5}    & \text{77.1}     & \text{78.9}  & \text{92.8}  \\
		 & \text{LORE$^*$\cite[]{Hangdi2023LORE}}  & \textbf{86.3} & \text{83.4}     & \textbf{84.8}  & \textbf{95.3}  \\
		\bottomrule
	\end{tabular*}
	\footnotetext{Best results are in \textbf{bold}. The "$^*$" indicates that the model first uses the layout analysis model to identify the table area, and then identifies the table structure separately in the table area. No "$^*$" means that the model directly recognizes the table structure of the entire PDF image.}
	\footnotetext{ }
\end{table}

\subsection{Experimental results}
\label{subsec:results}

\textbf{Wired table result} Table \ref{tab:result_main} presents the evaluation outcomes for the Chinese financial documents table dataset. We choose the LineCell model implemented in this paper and the latest LORE\cite{Hangdi2023LORE} model for comparison. Additionally, we investigated the impact of employing the layout analysis model for table area identification in the LORE\cite{Hangdi2023LORE} model. Analyzing the experimental results, we can find:

 \begin{table}[h]
    \caption{Compare with state-of-the-art methods on PubTabNet dataset.}
    \label{label:tsr_pubtabnet}
	\begin{tabular}{l|c|c|c|c|c}
		\toprule
	Methods & \makecell{Acc \\ (\%)} & \makecell{TEDS \\ (\%)} & \makecell{TEDS-Struct \\ (\%)} & \makecell{Inference \\time(ms)} &  \makecell{Model \\ Size(M)}\\
	\hline
	TableMaster\cite{ye2021TableMaster} & 77.90 &	96.12 &	- &	2144 &	253 \\
	\text{TableMaster$^*$\cite{ye2021TableMaster}} & 78.60 & - & 97.56 & 2764 &	260 \\
	\hline
	LGPMA\cite{qiao2021lgpma}  & 65.74 & 94.70 &	96.70 &	- &	177 \\
	\text{LGPMA$^*$\cite{qiao2021lgpma}} & 65.30 & - & 96.68 &	345 &	177 \\
	\hline
	SLANet\cite{ppstructurev2_li2022pp} & 76.31 &	95.89 &	97.01 &	766 &	9.2 \\
	\text{SLANet$^*$\cite{ppstructurev2_li2022pp}} & 76.03 & - & 97.33  &	798 &	9.2 \\
	\hline
	MTL-TabNet\cite{visapp23MTLTabNet} & - &	96.67 &	97.88 &	- &	289 \\
	\text{MTL-TabNet$^*$\cite{visapp23MTLTabNet}} & 79.10 &	- &	98.48 &	4520 &	289 \\
	\bottomrule
	\end{tabular}
	\footnotetext{ "$^*$" denotes the results of our assessment. Model size refers to the actual physical size of the model. Regarding the inference time of some models, we quote from the SLANet\cite{ppstructurev2_li2022pp} paper.}

\end{table}
	
The evaluation metrics exhibit superior performance on digital PDFs compared to image-based PDFs, with the F1-score demonstrating an 11.2\% increase on digital PDFs. This suggests that table extraction is more challenging in image-based PDFs, highlighting substantial room for improvement. The LineCell model outperforms the LORE\cite{Hangdi2023LORE} model by 4.2\% in F1-score on digital PDFs, while registering a 0.6\% lower F1-score on image-based PDFs. This indicates that the traditional LineCell model still has certain advantages in identifying wired tables within PDFs. Notably, the LineCell model achieves an F1-score of 98.4\% on digital PDFs and 84.2\% on image-based PDFs, showcasing its effective identification of wired tables in PDF documents.

By comparing the results of whether the LORE\cite{Hangdi2023LORE} model first uses the layout analysis model, it is evident that employing the layout analysis model before table structure recognition enhances the F1 score by 5.5\% and the TEDS-Struct score by 1.85\%. This underscores the effectiveness of incorporating the layout analysis model for predictive processing, resulting in a notable improvement in table recognition accuracy.

\textbf{Wireless table result} The PdfTable toolkit incorporates diverse models for wireless table structure recognition. To assess the accuracy of algorithm integration, we conducted evaluations on the \cite{zhong2020PubtabNet} dataset. The experimental results are shown in Table \ref{label:tsr_pubtabnet}. It can be found from the experimental results:

By comparing the Acc and TEDS-Struct metric of the four models, the maximum difference between our evaluation results and the original paper results is 0.7\%, falling within the acceptable margin of error. This preliminary validation underscores the accuracy of the algorithm integration.

From the perspective of inference speed and Acc metric, SLANet\cite{ppstructurev2_li2022pp} exhibits distinct advantages compared to other models. It achieves an Acc metric of 76\% with an average inference time of 798 ms. TableMaster\cite{ye2021TableMaster} and MTL-TabNet\cite{visapp23MTLTabNet} can attain higher Acc, their average inference times are considerably slower. Notably, the MTL-TabNet\cite{visapp23MTLTabNet} model achieves the best results, but the average inference time is as high as 4520 ms.

The LORE\cite{Hangdi2023LORE} model can also support wireless table recognition, and the TEDS metric reaches 98.1\% on the PubTabNet\cite{zhong2020PubtabNet} dataset. However, there are currently problems with the integration in PdfTable, and the experimental results have not been entirely replicated. Future optimizations are planned.

\subsection{Qualitative Assessment}
\label{subsec:example}

\begin{figure}[htbp]
	\subfigure[Original table image 1] {\includegraphics[width=.3\textwidth]{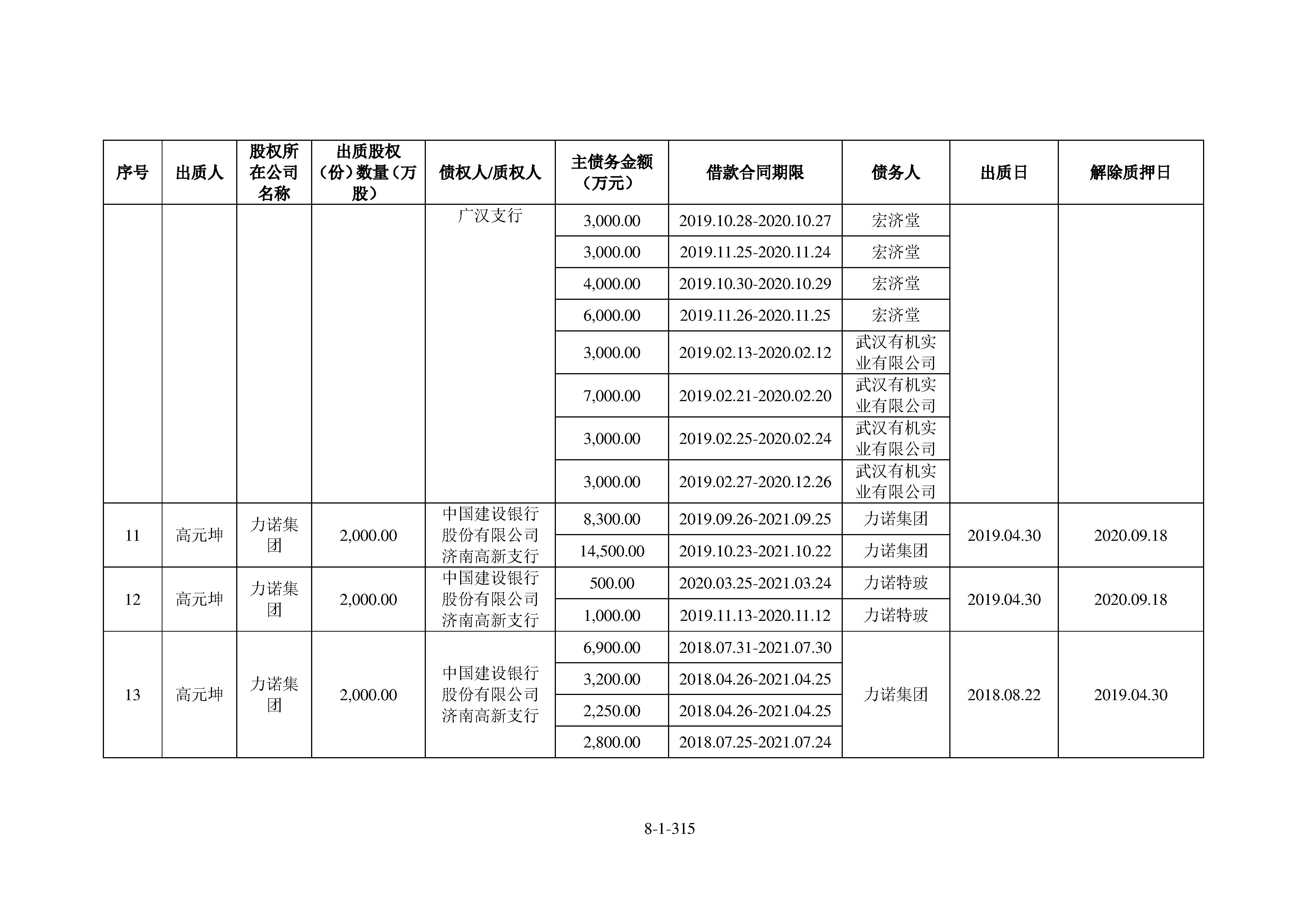}}
	\subfigure[LineCell prediction results 1] {\includegraphics[width=.3\textwidth]{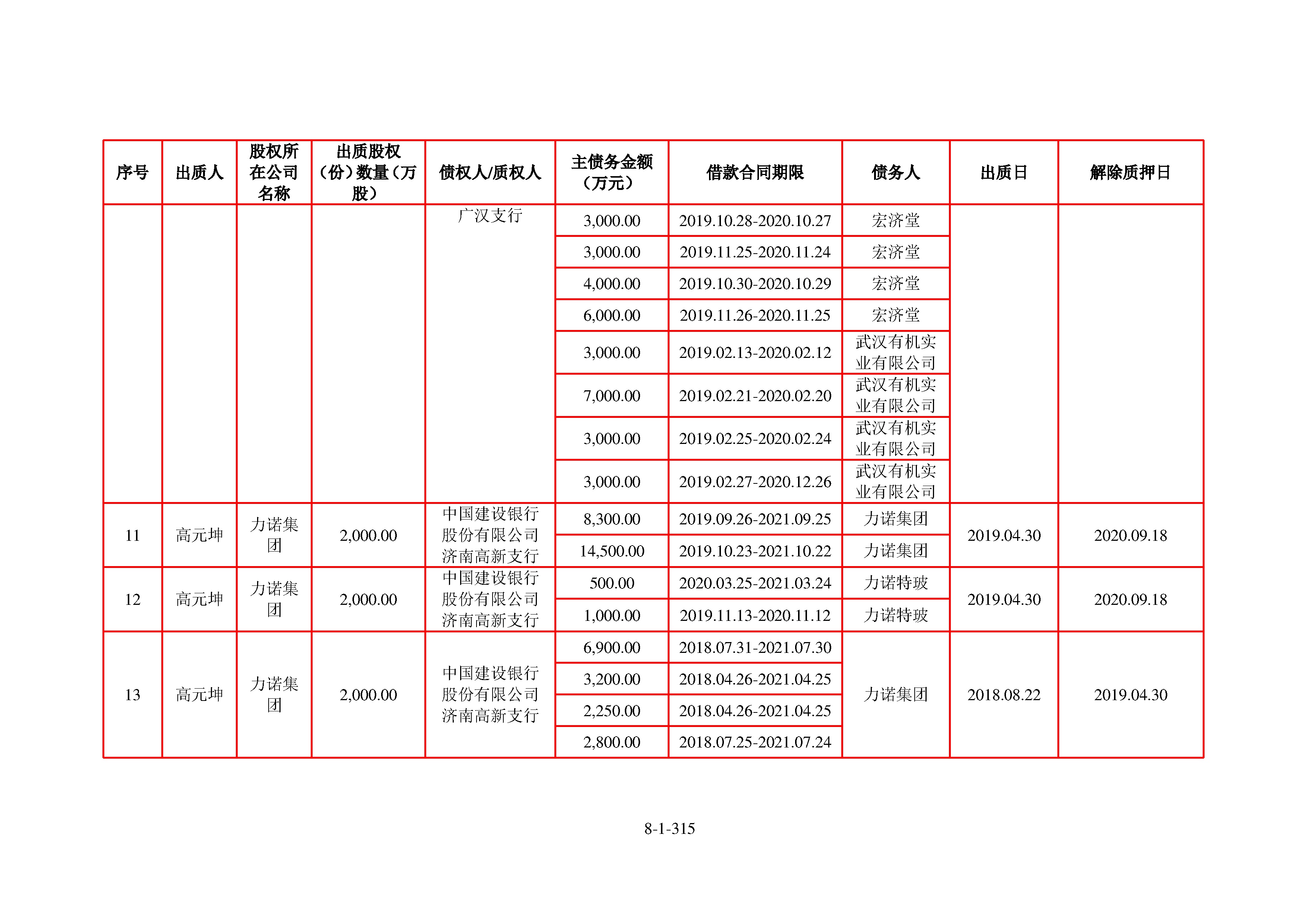}}
	\subfigure[LORE prediction results 1] {\includegraphics[width=.3\textwidth]{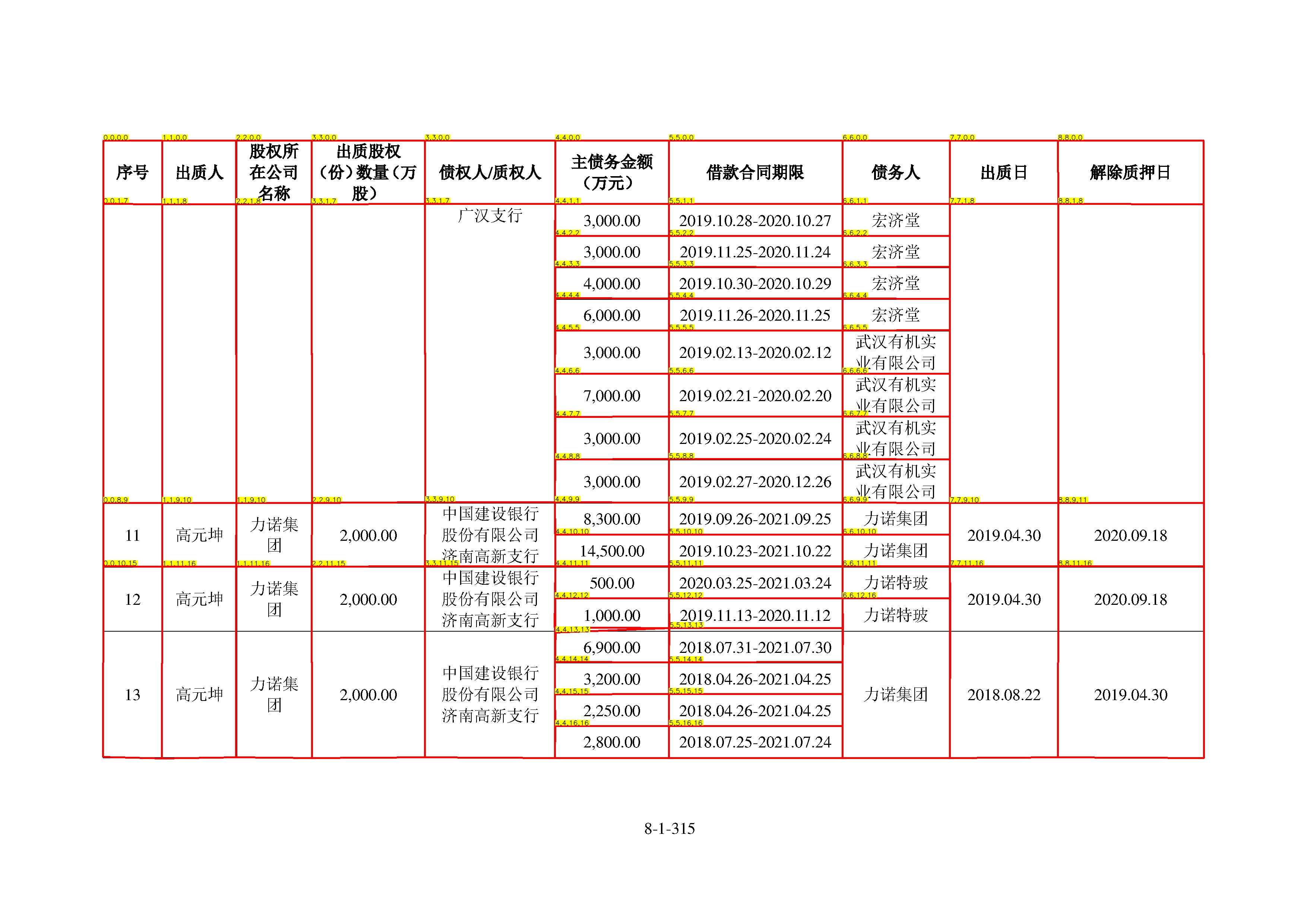}}
    \\
    \subfigure[Original table image 2] {\includegraphics[width=.3\textwidth]{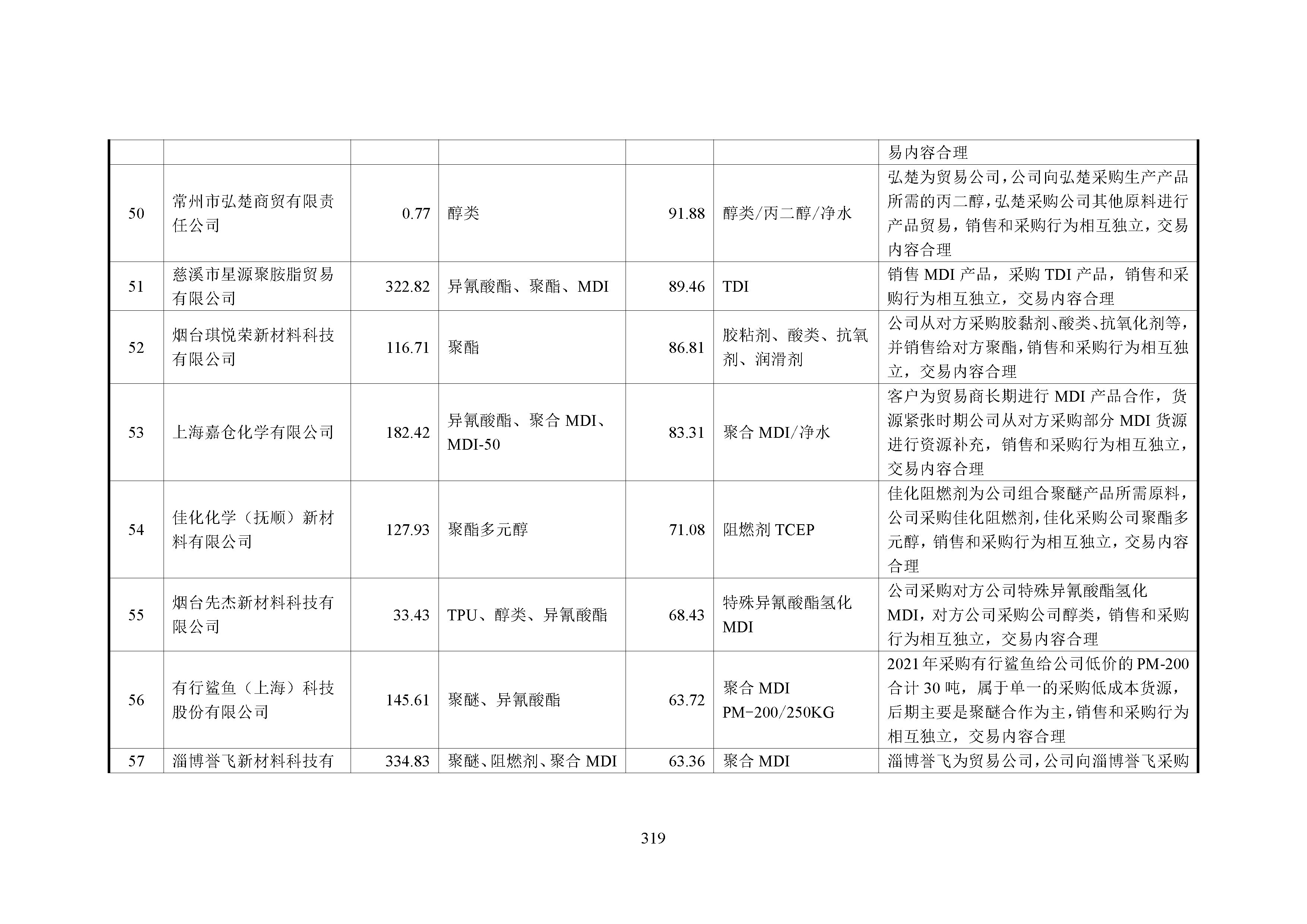}}
    \subfigure[LineCell prediction results 2] {\includegraphics[width=.3\textwidth]{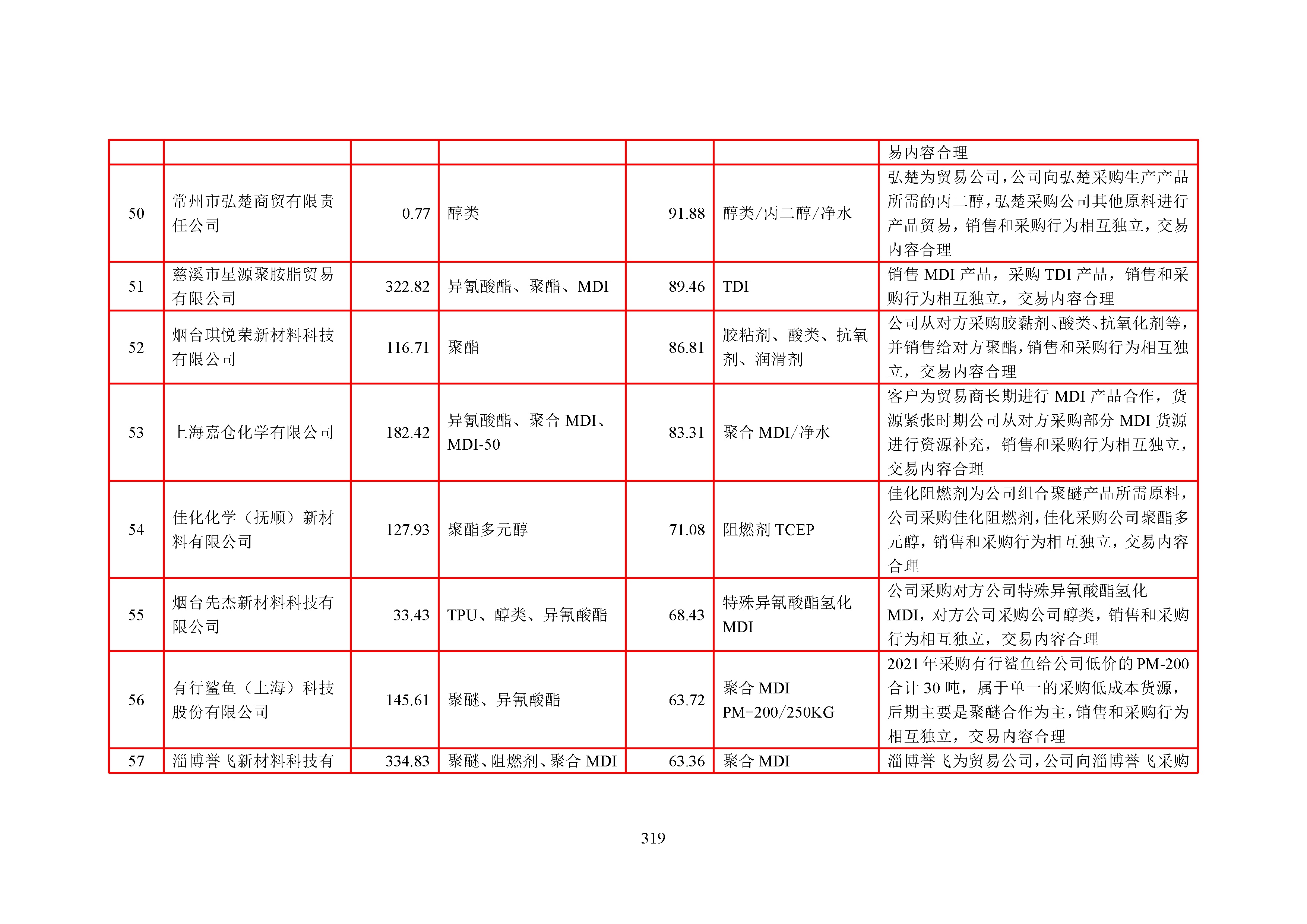}}
    \subfigure[LORE prediction results 2] {\includegraphics[width=.3\textwidth]{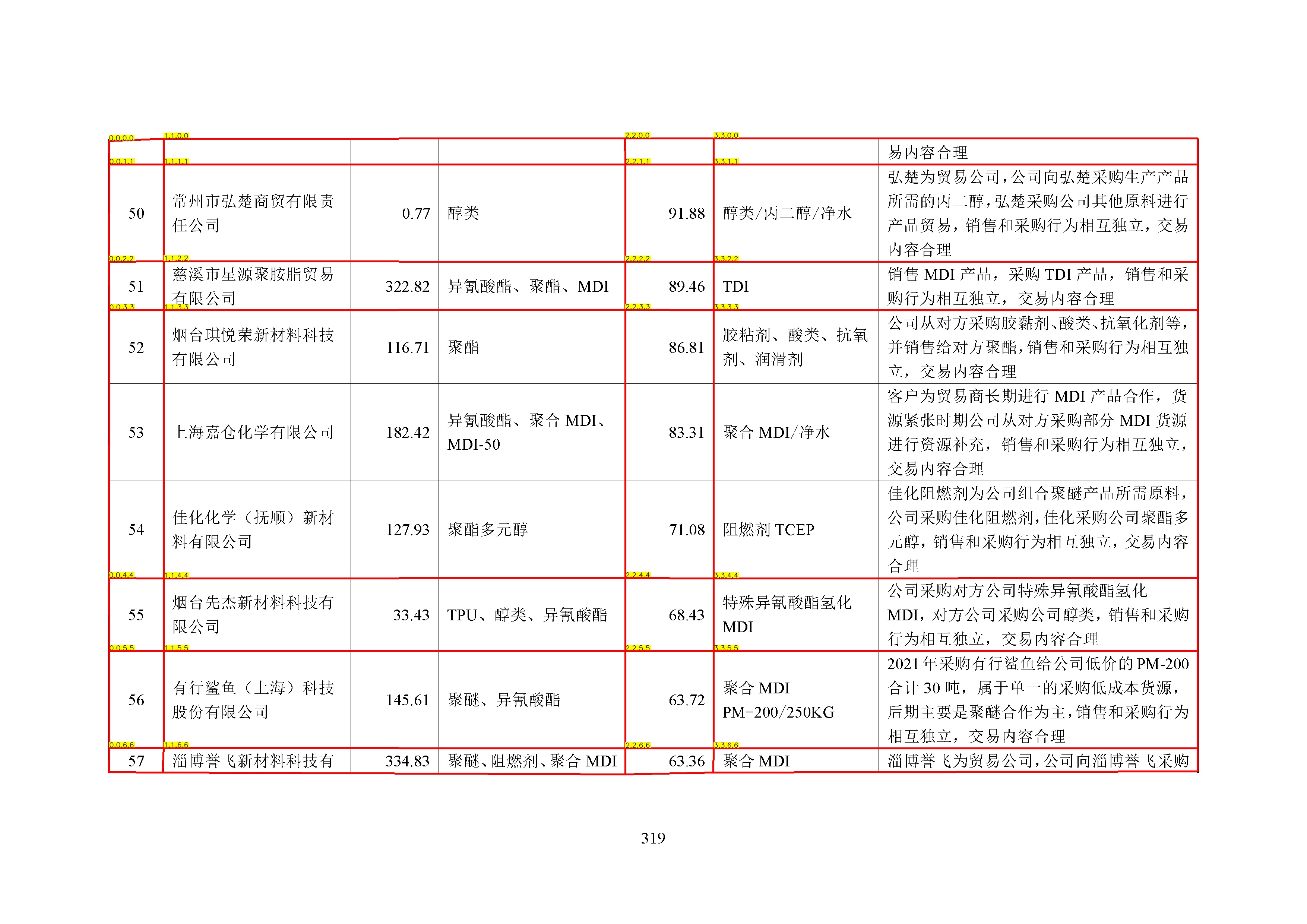}}
	\caption{Qualitative results of LineCell and LORE on digital PDF. The red border represents the identified cell.}
    \label{figure:example}
\end{figure}

The qualitative results in Figure \ref{figure:example} show that in some cases, the LORE\cite{Hangdi2023LORE} model predicts that some cells in the table cannot be accurately identified, whereas LineCell can accurately identify all cells.

\section{Conclusion and future work}
\label{sec:conclusion}
In this paper, we introduce a novel end-to-end PDF table extraction toolkit, PdfTable, designed for seamless table extraction from both digital and image-based PDFs. The toolkit integrates various existing models, including those for layout analysis, table structure recognition, OCR detection, and OCR recognition. This integration allows for flexible combinations to adapt to diverse application scenarios. To validate the efficacy of the PdfTable toolkit, we annotated a small dataset of wired tables. Concurrently, we evaluated the wireless table recognition model on the PubTabNet\cite{zhong2020PubtabNet} dataset, confirming the accuracy of the algorithm integration. In the future, we will optimize this toolkit from the following aspects: 1. Developing new algorithms to differentiate wired tables from wireless tables; 2. Incorporating the ability to fine-tune integrated models, such as table recognition models; 3. Enhancing the toolkit's capacity to recognize wired tables in image-based PDFs.

\section{Declarations}

\textbf{Conflict of interest} The authors have no conflicts of interest to declare that are relevant to the content of this article.

\textbf{Ethics approval} This article has never been submitted to more than one journal for simultaneous consideration. This article is original.



\textbf{Data Availability} The datasets analysed during the current study are available in the \url{https://github.com/ibm-aur-nlp/PubTabNet}. 

\textbf{Code availability} Code and data used in this paper are publicly available at \url{https://github.com/CycloneBoy/pdf_table}.


\bibliography{pdf_table}

\end{document}